\documentclass[sigconf]{acmart}

\usepackage{booktabs} 

\usepackage{amsmath}    
\usepackage{mathtools}  
\usepackage{amssymb}    
\usepackage{verbatim}   
\usepackage{color}      
\usepackage{hyperref}   
\usepackage{multicol}  %
\usepackage{multirow}
\usepackage{ifthen} 

\usepackage{graphicx}
\usepackage{caption}
\usepackage{subcaption}

\usepackage{lipsum}       
\usepackage{changepage}   
\usepackage{enumitem}

\usepackage{soul}
\usepackage{balance}
 
\definecolor{mypink}{rgb}{0.858, 0.188, 0.478}
\definecolor{mygray}{gray}{0.6}

\usepackage[normalem]{ulem}

\setcopyright{rightsretained}

\copyrightyear{2017}
\acmYear{2017}
\setcopyright{acmcopyright}
\acmConference{KDD '17}{}{August 13-17, 2017, Halifax, NS, Canada}
\acmPrice{15.00}
\acmDOI{http://dx.doi.org/10.1145/3097983.3098078}
\acmISBN{978-1-4503-4887-4/17/08}

\begin{document}
\title{A Data Science Approach to Understanding Residential Water Contamination in Flint}

\author{Alex Chojnacki}
\authornote{The six student authors are alphabetically ordered first, followed by the two faculty authors, also alphabetically ordered.}
\affiliation{%
  \institution{University of Michigan}
}
\email{thealex@umich.edu}

\author{Chengyu Dai}
\affiliation{%
  \institution{University of Michigan}
}
\email{daich@umich.edu}

\author{Arya Farahi}
\affiliation{%
  \institution{University of Michigan}
}
\email{aryaf@umich.edu}

\author{Guangsha Shi}
\affiliation{
  \institution{University of Michigan}
}
\email{guangsha@umich.edu}

\author{Jared Webb}
\affiliation{%
  \institution{Brigham Young University}
}
\email{webb@mathematics.byu.edu}

\author{Daniel T. Zhang}
\affiliation{%
  \institution{University of Michigan}
}
\email{dtzhang@umich.edu}

\author{Jacob Abernethy}
\affiliation{\institution{University of Michigan}}
\email{jabernet@umich.edu}

\author{Eric Schwartz}
\affiliation{\institution{University of Michigan}}
\email{ericmsch@umich.edu}

\renewcommand{\shortauthors}{A. Chojnacki et al.}

\begin{abstract}
When the residents of Flint learned that lead had contaminated their water system, the local government made water-testing kits available to them free of charge. The city government published the results of these tests, creating a valuable dataset that is key to understanding the causes and extent of the lead contamination event in Flint.  This is the nation's largest dataset on lead in a municipal water system.

In this paper, we predict the lead contamination for each household's water supply, and we study several related aspects of Flint's water troubles, many of which generalize well beyond this one city. For example, we show that elevated lead risks can be (weakly) predicted from observable home attributes. Then we explore the factors associated with elevated lead. These risk assessments were developed in part via a crowd sourced prediction challenge at the University of Michigan.  To inform Flint residents of these assessments, they have been incorporated into a web and mobile application funded by \texttt{Google.org}. We also explore questions of self-selection in the residential testing program, examining which factors are linked to when and how frequently residents voluntarily sample their water. 
\end{abstract}

%
%



\maketitle

\section{Introduction}
We now understand the Flint Water Crisis as a disaster with many facets: environmental, socio-economic, political, and infrastructural, among others. The dire problems affecting the city's water started in April 2014 when, as a short-term cost-saving measure, city officials opted to switch the water supply from Lake Huron to the Flint River. Not long after the switch, residents began to notice an unpleasant odor and discoloration in the water flowing from their taps.  While water testing data reported by state government officials passed regulations from the U.S. Environmental Protection Agency (EPA), data collected by outside academics from Virginia Tech suggested otherwise. This independent academic work found water lead levels dramatically higher than the threshold  allowed by the EPA's Lead And Copper Rule. It was not until September 2015, following a report by a pediatrician observing a dramatic rise in lead levels\footnote{It is now well established that lead-contaminated water poses significant health risks, particularly for children \cite{Toronto2014}} in the blood of Flint children \cite{hanna:2016}, that the water crisis began to receive serious attention from government officials. In December 2015, Flint's mayor declared a state of emergency, and agents from both the Michigan Department of Environmental Quality (DEQ) and the EPA embarked on thorough investigations. By late 2015 and early 2016, the media had elevated the Flint Water Crisis into a major national and international news story.

Eventually, the immediate cause was understood: the water from the Flint River was significantly more corrosive than local officials had thought. This, and other governmental failures, resulted in improper water treatment. Central to the problem was that, like many U.S. cities, Flint's water infrastructure contains tens of thousands of lead pipes. These pipes typically are treated with beneficial chemicals to develop thick layers of deposits, which protect water against contamination from heavy metals. Treated incorrectly, however, Flint's corrosive water began to erode these protective layers and ultimately, lead particles leeched from the pipes into the city's drinking water. Though Flint returned to Lake Huron's water supply in October 2015, the damage was done, and pervasive lead contamination continued to be detected through 2016. While the EPA determined the water was safe to drink with a filter by mid 2016, many issues remain and citizens continue to rely on bottled water \cite{detroitnews:Flint}. The city's most vulnerable residents, namely children, pregnant women, and the elderly, have likely been exposed to lead in the water, and many questions about the lasting impact remain unanswered.

As Flint's water crisis has continued to unfold, affecting as many as 35,000 homes, both city and state officials have been faced with daunting questions: what is the best way to direct scarce resources? How can bottled water and water filter technology be efficiently distributed?  Where should volunteers be sent to educate residents? As the city has embarked on a highly expensive pipe removal program, where a replacing a single home's water service line can cost around \$5,000, officials have asked the obvious question: which homes are most at risk for lead contamination? Flint's recovery depends greatly on isolating which properties are most in need of attention. This question is important beyond Flint, as other cities and towns with aging infrastructure continue to address lead and other heavy metal abatement.

In the present paper, we consider the problem of estimating the risk of lead contamination in home drinking water. This work relies on a large collection of water samples taken by residents and government officials throughout the crisis. Beginning in late 2015, the State of Michigan initiated program allowing any resident to submit a tap water sample for testing. This dataset is a publicly available collection of over 25,000 tests, and it provides a glimpse into the causes and extent of water lead contamination in Flint; it is indeed the largest dataset collected on lead in a municipal water system. We combine these measurements with several other data sources, including census data, property attributes, geographical information, and infrastructure records, and we use the combined data to answer several statistical and analytical questions.  Among these are:
\begin{itemize}
  \item \emph{To what extent can we predict elevated lead in a home's drinking water?}
  \item \emph{What attributes of a home are associated with lead contamination?}
  \item \emph{How can we address the sampling bias of volunteer residential testing?}
\end{itemize}
We present a number of additional results, and we conjecture that many of these observations will generalize beyond Flint.

\subsection*{Flint's Water Contamination: A Birds-Eye View}

Before we begin our analysis, let us give an overview of the lead testing data and a brief analysis. When a resident takes a water sample and submits this water sample for testing, the state determines the lead content (typically by mass spectrometer) and reports the result in \emph{parts per billion} (ppb). The data released by the state rounded these values down to the nearest integer. Thus when we say that a sample had ``no detectable lead'' we mean less than 1 ppb. It is important to note that, despite what one may infer from headlines, nearly half of all homes had no detectable lead, and around 80\% of measurements from the residential testing program were below 5 ppb.

These lead levels still warranted attention according to the law. The US Congress passed what is known as the Safe Drinking Water Act in 1986, which instructed the EPA to develop regulations limiting heavy metals in drinking water. Pursuant to the act, the EPA developed what is now known as the Lead and Copper Rule (LCR), issued in 1991, requiring municipal water utilities to enforce a set of guidelines for allowable levels of lead and copper. More precisely, the LCR requires that at regular intervals a municipality must take a set of water samples from a range of properties, and that the 90$^{\rm th}$ percentile lead measurement must fall below 15 ppb. As a result of these EPA requirements, throughout the paper we emphasize this 15 ppb threshold.

It is worth noting that, from the perspective of public health, this value of 15 ppb is rather arbitrary. It is very challenging to determine precisely the risks to human health from lead contamination in water, and most epidemiological work aimed at understanding adverse effects from consuming dissolved lead can provide only coarse answers \cite{ngueta2016use}; public health experts typically say that ``no level of lead is safe.'' The current guidelines should be viewed only as a workable regulatory framework.

\begin{figure}[h]
\centering
\includegraphics[width=8cm]{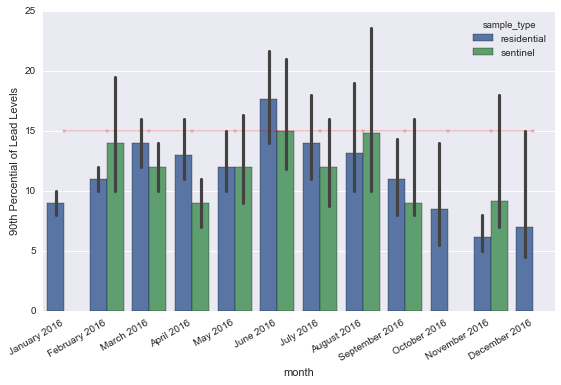}
\caption{Comparing the 90$^{\rm th}$ percentile of lead readings, on the sentinel data vs. the voluntary residential testing data.}
\label{fig:nintieth}
\end{figure}

Based on this law, the key quantity is the estimate of the 90$^{\rm th}$ percentile of lead readings. We describe this quantity in Figure~\ref{fig:nintieth} for each month of 2016, drawing from both the government-run sentinel program and the larger voluntary residential testing program. Using data from the state's sentinel program, we found during a period in February only between 8 and 15 percent of homes had lead above the federal action level of 15 ppb. Lead measurements are confounded by weather and temperature, which is likely the reason behind the summer rise in lead levels. But in general it is hard to draw simple conclusions about the trend of lead contamination in Flint.

\begin{figure}[h]
\centering
\includegraphics[width=5cm]{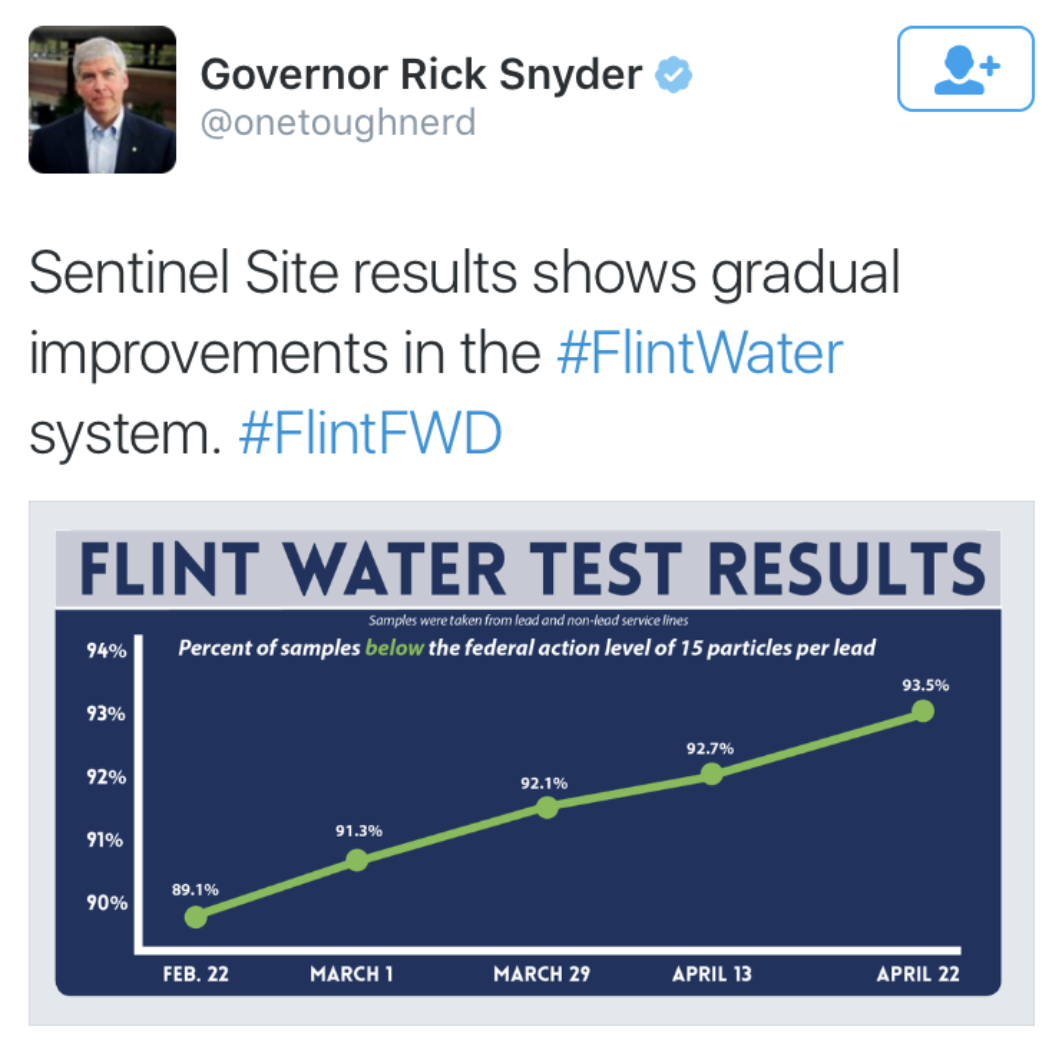}
\caption{Governor Rick Snyder announcing improved Flint water testing results on April 22, 2016. The tweet was quickly deleted, and results then worsened over the summer. Note that the plotted points of the line do not correspond to the y-axis labels, and x-axis is not linear in time.}
\label{fig:sneakysnyder}
\end{figure}

Despite the statistical issues, a result of these guidelines has been significant political attention paid to what percent of homes that test at or below the 15 ppb threshold. This was especially true in Flint where it was alleged that government officials manipulated data to achieve compliance. At the height of the political firestorm, the embattled Governor Rick Snyder put out a tweet (seen in Figure~\ref{fig:sneakysnyder}) celebrating good news about the elevated lead levels. The tweet was deleted within a day, but we were able to grab a screenshot. Our own analysis, as displayed in Figure~\ref{fig:nintieth}, however, rejects the conclusion of Governor Snyder's deleted tweet about the distribution of lead levels over time.

\paragraph{Related Work} 
Much of the work up until this point was conducted by Marc Edwards' team from Virginia Tech, who independently monitored of lead water levels \footnote{http://www.flintwaterstudy.org}. Their efforts have helped raise awareness and reveal the severity of the problem. In addition, \cite{baum:2016} provides an overview of the water crisis and discusses strategies for risk management in Flint. Further, there is some work analyzing some similar trends that we observe in lead levels over time \cite{Goovaerts2017temporaltrends,goovaerts2017monitoring}. But to the best of our knowledge, we are the first to apply predictive modeling techniques to help with the Flint Water Crisis.\footnote{The authors recognize some of their own work has been presented elsewhere \cite{abernethy2016flint}.}

\section{Data}

This paper incorporates a diverse range of datasets related to properties in the city of Flint. One of the main contributions of our work is acquiring and merging these datasets into a single dataset. Some of these datasets are publicly available from the state of Michigan, and others were provided by the city and other sources at our request, as noted. We detail each dataset. 

\subsection{Residential Water Testing}

\begin{figure}[h]
\centering
\vspace{0.07in}
\includegraphics[width=0.45\textwidth]{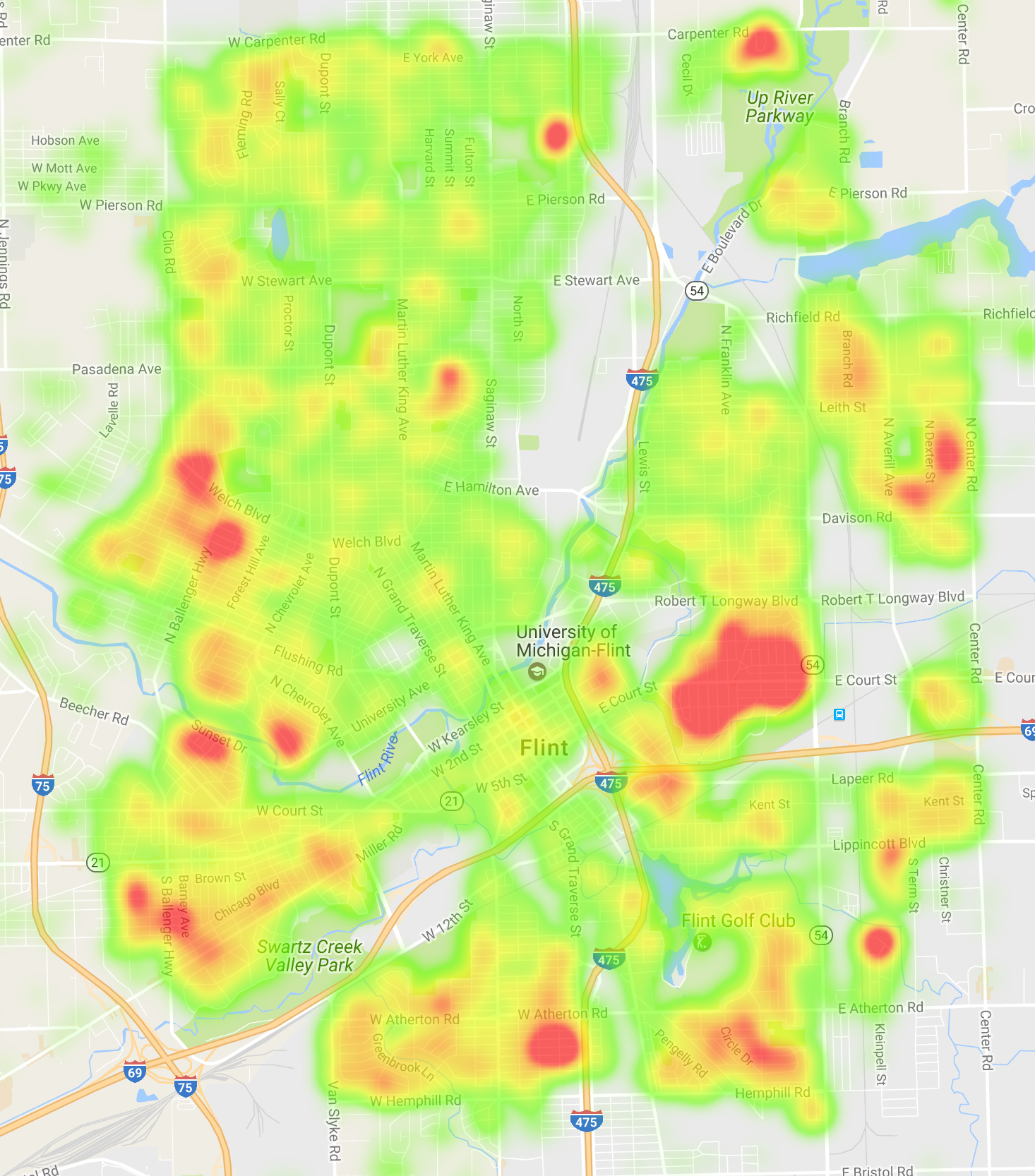}
\caption{Locations of voluntary residential water tests in Flint. Color corresponds to the level of lead contamination (parts per billion). We observe that elevated lead readings are highly geographically diverse.}
\label{fig:heatmap}
\end{figure}
The vast majority of the lead water level data in Flint comes from water samples submitted voluntarily by residents. The city of Flint provides free water testing services to all of its residents, who are able to pick up testing kits from a local distribution center. Residents then collect water from their own homes and submit the samples to be analyzed by the Michigan Department of Environmental Quality. Since this program began in September 2015, over 25,000 tests have been conducted from 15,000 unique locations (as of May 2017).  The results are available on the State's website \footnote{\url{http://www.michigan.gov/flintwater/}}. For each sample we are given the date the sample was submitted, the lead and copper levels, and the address of the residence. In Figure ~\ref{fig:heatmap}, we show the locations and lead readings for these tests. Measuring lead contamination is a highly noisy process, and even repeated measurements at the same source produce highly variable results. We can observe this directly in the data because a subset of homes had their water tested on multiple occasions. \footnote{In Section \ref{sec:behavior}, we address some of the reasons for residents testing their water more than once.} The correlation in (log) lead levels between first and second samples is modest (Pearson correlation coefficient 0.465 for voluntary residential testing and 0.522 for the sentinel program.

This noisy measure has an effect on performance of our predictions, as we will see later.  There are many causes for this noise, but one major source is the delicate nature of sampling a home's water. Residents are asked to sample the first liter of water from their tap first thing in the morning, with the hope of getting water that has been stagnant in the plumbing, but a toilet flush or running the shower can significantly affect the concentration of various contaminants.

\subsection{Sentinel Water Testing}
\label{sec:sentinel}

As news of the crisis broke, Michigan DEQ initiated what is called the ``sentinel program,'' in which over 400 homes were selected to be tested multiple times over many months. These were homes that were considered to be especially at risk of lead contamination---many were known to have a lead service line, for example---and they were drawn from diverse neighborhoods around the city. These sites were chosen to be a representative sample, and the state received some guidance from other academics for selecting these homes.

Data from the sentinel program has been made publicly available at \url{http://michigan.gov/flintwater}.\footnote{The sentinel data omit the full addresses of the homes, but our team was able to get access to these records with help from the Michigan Governor's office. This allowed us to link each home to the many variables describing each parcel of property.} 

One of the challenges with determining lead contamination levels is determining which homes to test. The EPA requires water systems to select homes that are at greater risk of elevated lead in their tap water, according to the Lead and Copper Rule, but this leaves much to the discretion of officials who can seek data points in order to produce more optimistic (or pessimistic) estimates. Indeed, investigators have questioned the selection of homes in Flint, for instance some were in a more newly-developed neighborhood \cite{NYT:Flint,goovaerts2017monitoring}.

Sentinel sites were visited for water tests a varying number of times, with some homes tested fewer than 5 times, while others were tested more than 10 times. The samples were taken at roughly weekly intervals, early in 2016, and then less frequently as the year went on. While the sentinel data represents a smaller set of homes than the voluntary residential testing program, we generally assume the sentinel data to be much more reliable as the residents in these homes are given more  direct instructions, by workers and other officials, on how to correctly take a water sample. The bottles are picked up by DEQ officials and others for chemical testing.

\subsection{Parcel data}

The city provided us with detailed records of the 55,893 parcels of land in Flint. This data contains information on the property's age, location, and value, in addition to other characteristics. This data is not publicly available online in this exact form, but a very similar dataset is freely in an ARCGIS format, known as Flint 2014 Housing Data.\footnote{The parcel data is made available online by Housing and Urban Development and contains information taxpayer, ownership, land use, and vacancy and is accessible here:   \url{https://www.arcgis.com/home/item.html?id=bcd87aa254d34ae6b66475beaf17d59a\#overview.}
We initially received a version of this parcel data directly from the City of Flint.} We used the Google Maps API to merge noisy address data.\footnote{Thanks to a grant and API access from Google.org.} Those samples that did not correspond to Flint parcels were discarded. After merging and discarding non-Flint parcels, 55,857 parcels remained in our dataset. 

The key step was merging the parcel data with the lead testing data. We matched the address of each lead test to the address of the corresponding parcel of land in the city records. Because a parcel can contain multiple residences and residents are free to submit as many tests as they would like, we often have multiple tests that correspond to a single parcel. On the other hand, because many properties in Flint are vacant and residents are not required to submit tests, most parcels have no associated lead test. 

\begin{figure}[h]
\centering
\includegraphics[width=0.45\textwidth]{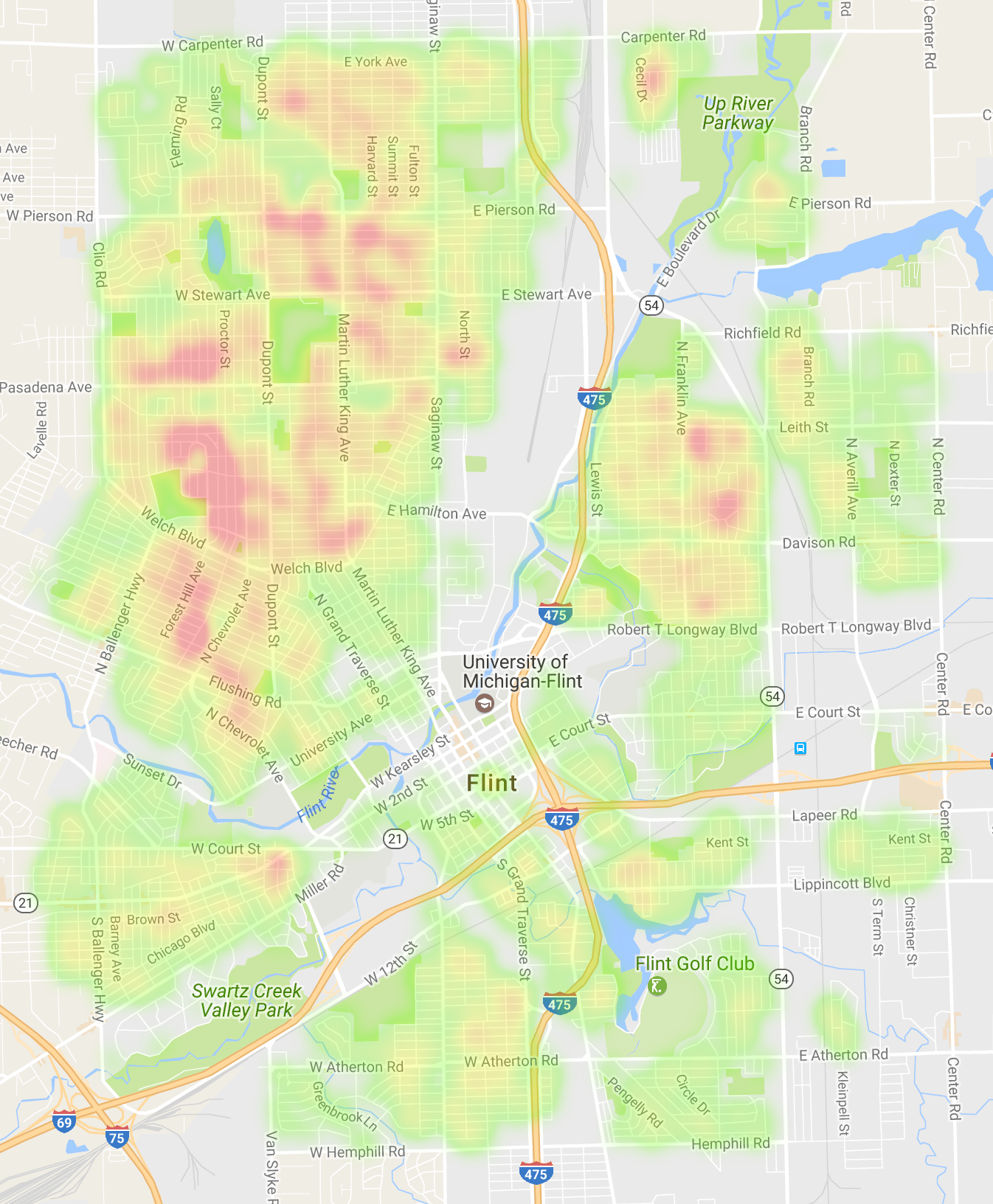}
\caption{There are many abandoned homes in Flint MI. This heatmap displays the density of (likely) unoccupied properties.}\label{fig:vacancies}
\end{figure}

An important challenge working with residential data on Flint is a striking fact: \emph{Flint has the highest rate of vacant homes in any municipality across the US} \cite{ResidentialPropertyVacancyRate}. Figure \ref{fig:vacancies} shows the density map of vacant homes in on the Flint map.  We have two variables serving as weak signals of occupancy: does the home has an active U.S. Postal Service account, and was the 2014 Housing Condition survey. In our discussions to follow, in Section \ref{sec:behavior}, we carefully consider vacancy, and characterize the a household's decision to submit a residential water test along with  whether that test will have an elevated lead reading.

\subsection{Service line data} \label{sec:serviceLines}
Water service lines are the pipes that connect each property in Flint to the water distribution system, often called the ``water main''. A home's water service line is typically composed of two different segments: public and private. The \emph{public} service line which is the pipe connecting the water main to the property ``curb box'', which is an underground device owned by the municipality that contains a shutoff valve.  The \emph{private} service line connects the curb box through front lawn and runs into the home's water meter.

Service lines can be made out of any number of materials, including lead, copper, galvanized steel, plastic, and other metal alloys. Unfortunately, there is not a definitive record of the service line material for every home. Initially, the City of Flint struggled to produce any service line records. Eventually they discovered a set of 45,000 $3"\times 5"$ index cards and a set of municipal maps from the water department with handwritten annotations \cite{NPR:URL}. The information in these maps was painstakingly digitized by a group of students at the University of Michigan, Flint, GIS center. This project was spearheaded by Dr. Marty Kaufman, the faculty director of the center.  It was noted that the city records are not always accurate and reliable. For more details about the service lines see \cite{webb2006flint}.\footnote{The replacement of lead and galvanized service lines became a top priority for the City of Flint in February 2016. By May 2017, over \$100 million in State and Federal funds had been appropriated for Flint service line replacement, managed by by the Flint Fast Action and Sustainability Team.}

\subsection{Census Block Level Data}

The previous data tell us much about the physical properties of the homes in Flint, but they do not tell us much about the people that live in them. They also provide a richer understanding of the affected populations. The census conducted by the U.S. Census Bureau has precise, parcel-level demographic data, but this data is not made available until many years after it is gathered to protect citizens' privacy. The American Community Survey (ACS), however, is a survey conducted by the U.S. Census Bureau that supplements their census data with demographic and economic data. The results are provided at the level of census block groups.

Using the American Fact Finder website\footnote{\url{https://factfinder.census.gov}}, we acquired data about race, age, family structure, languages spoken, household income and rent values for each block in Flint city limits.  The parcel data includes census tract, block group, and block information for each parcel, so these block-level census data were merged with the other parcel-level data.

\section{Predicting Lead Levels}

\begin{figure}[h]
\centering
\includegraphics[width=0.48\textwidth]{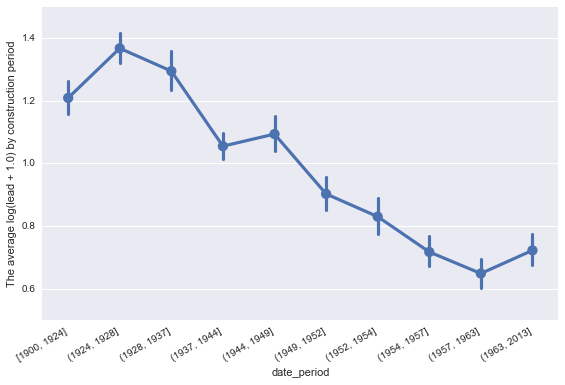}
\caption{When averaged over many parcels, lead levels display a number of very clear trends. Homes built after in the 1950s and later display significantly lower lead levels than homes built in the early 1900s.}
\label{fig:loglead_by_year}
\end{figure}

In the present section, we present our predictive models of water lead levels, allowing us to understand the factors related to high lead risk and to provide predictions for homes that had not yet been tested. In the previous section, we discussed the challenges associated with lead testing data, particularly due to the noisy nature of the sampling process. But A closer look at lead level data from Flint provides a much more nuanced picture, A number of home features correlate quite strongly with elevated lead, and we note one example that should not come as a great surprise: \emph{the age of the property}. In Figure~\ref{fig:loglead_by_year} we report average log(lead levels +1) grouped by the year of construction for these homes, and the downward trend is quite stark.

Good lead risk predictions can inform public health policy in Flint. They can also provide insight into what factors are producing contaminated water.  In this section we discuss classification models that predict whether a water sample submission will test above the EPA action level of 15 ppb.

\subsection{Model Selection and Optimization}

To create our training data, we join the residential volunteer data with the merged parcel data so that each sample has a corresponding parcel. Note that not every home in Flint has submitted a water sample to be tested.  Similarly, several homes have submitted many samples, and these will have a row in the training data for each individual test.

For each row in our dataset, there are 71 features, coming from the parcel dataset, service line dataset, and census dataset.  One-hot encoding is performed on all categorical features. The target variable is the binary classification of homes with water tests above 15 ppb and below 15 ppb.

Training sets contained 75\% of the samples while the test sets were assigned the remaining 25\%. We split the data carefully into training and test sets creating two non-overlapping sets of parcels to prevent data leakage stemming from parcels with multiple water tests. Hyper-parameters for each model were chosen via a grid search. Finally, the \texttt{calibration} module from the \texttt{scikit-learn} was used to calibrate the predicted probabilities of the classifier.

\begin{figure}
\centering
\includegraphics[width=0.4\textwidth]{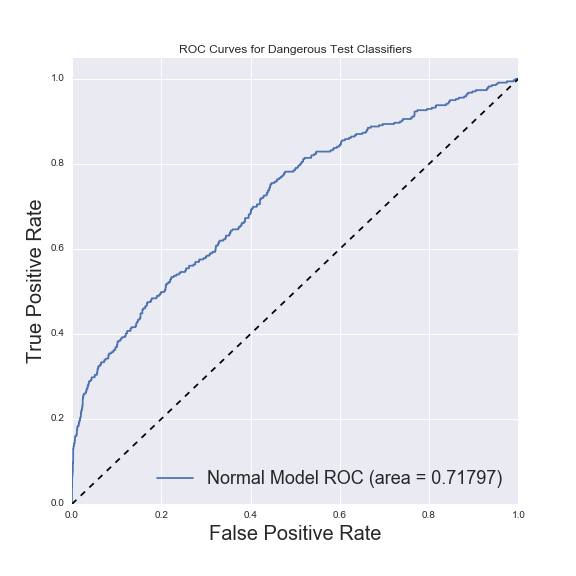}
\caption{ROC curve of a typical train/test split.}
\label{fig:lead_roc}
\end{figure}

We constructed various models using the $\texttt{scikit-learn}$ libraries and the $\texttt{XGBoost}$ python package \cite{chen2016xgboost}.  Tree based methods, such as random forests, performed the best, with the $\texttt{XGBoost}$ gradient boosted tree classifier achieving the best prediction result. The cross validation score after 250 runs for the classifier was $0.72 \pm 0.01$. A typical ROC curve is shown in Figure $\ref{fig:lead_roc}$.  The $\texttt{XGBoost}$ parameters are found in Table $\ref{tab:params}$.

\begin{table}
\centering
\caption{Grid search best parameters for $\texttt{XTBoost}$}
\label{tab:params}
\begin{tabular}{|c|c|}
\hline
Number of Trees          &  512 \\
\hline
Training Subsample Ratio &  0.9 \\
\hline
Tree Column Sample Ratio &  0.6 \\
\hline
Max Depth                &  3 \\
\hline
$\gamma$                 &  0.1 \\
\hline
$\alpha$                 &  0.5 \\
\hline
$\lambda$                &  1 \\
\hline 
\end{tabular}
\end{table}

\begin{figure}
\includegraphics[width=0.5\textwidth]{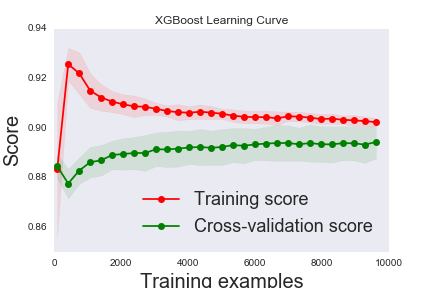}
\caption{The learning curve was produced using 10-fold cross validation and the $\texttt{scikit-learn model\_selection}$ module.  The convergence and small gap in the curve indicate that adding more data is unlikely to improve predictions.}
\label{fig:lead_learning}
\end{figure}

The learning curve for is shown in Figure $\ref{fig:lead_learning}$.
The convergence in the learning curve indicates that the model has been saturated with data. The initial steep decline in the training score indicates inherent bias in the model without sufficient data, but it declines with appropriate numbers of samples.

We also implemented various regression models, directly modeling the continuous non-negative value of lead levels (ppb).  However, compared to modeling the binary variable using the $15$ ppb threshold, these consistently produced inferior results.  For example, a collection typical \texttt{xgboost} regression models had a mean squared error of $305 \pm 72$.  When the predicted lead levels were converted into a $<15$ ppb classifier, AUC scores dropped to $0.63 \pm 0.1$.  This lackluster performance of the continuous regression model is likely driven by both the large range of target values and measurement error, high variance in lead levels even within the same parcel.  The perceived weakness of the regression models lead us to focus exclusively on classification.

\subsection{Results}

\begin{figure}
\centering
\includegraphics[width=0.5\textwidth]{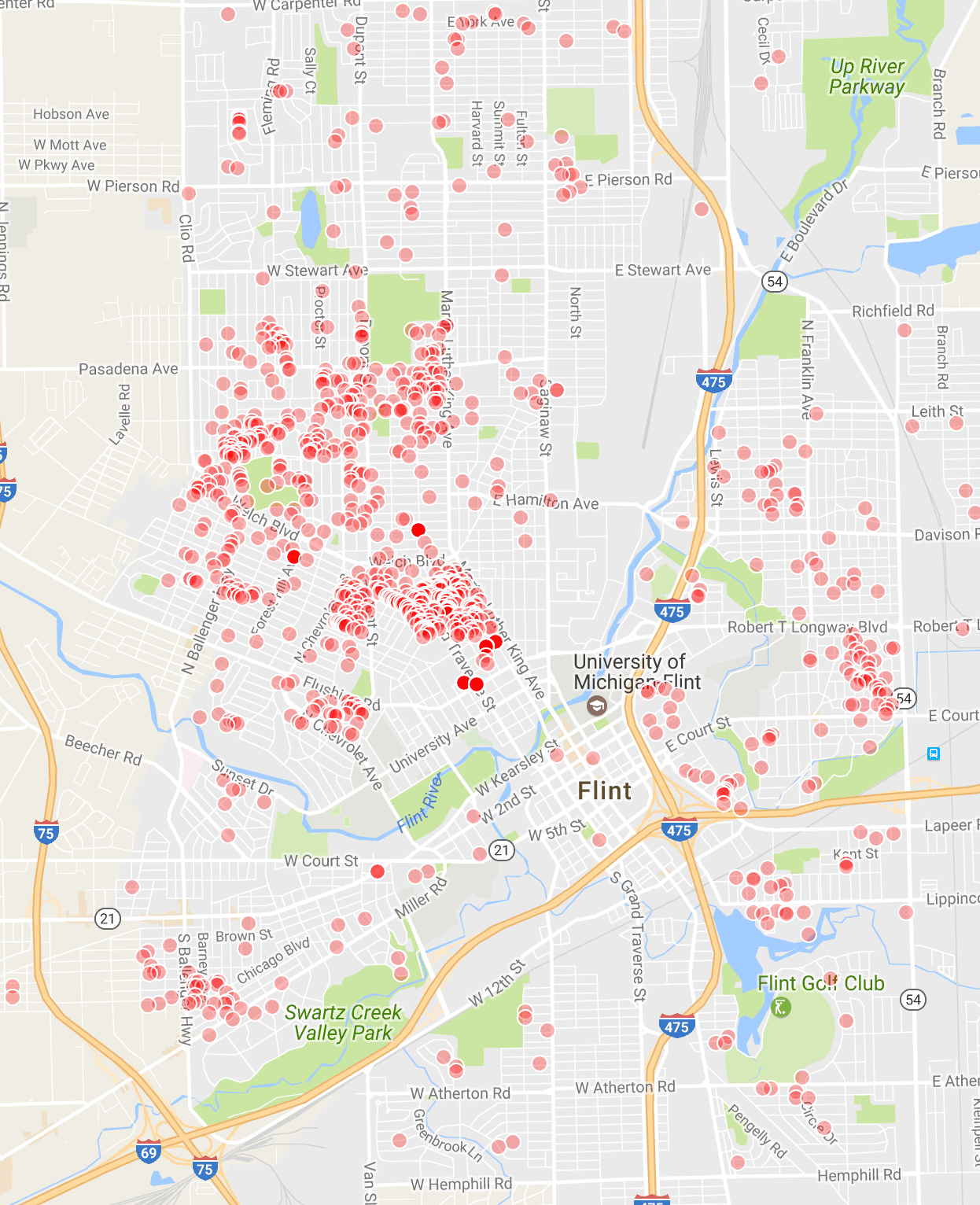}
\caption{The 1000 parcels with the highest probability to submit a water sample with lead above the EPA action level.}
\label{fig:lead_predict}
\end{figure}

After we determined the best model for predicting the water tests, we generated a prediction on all the parcels in the city of Flint. Figure \ref{fig:lead_predict} summarizes the location of 1,000 homes predicted to be most likely to have lead in their water which is above the EPA action level.  The homes in Figure \ref{fig:lead_predict} have not submitted lead tests yet. These predictions serve an important purpose, as they provide a risk assessment for homes that were never tested during the peak of the crisis. The analysis provides a predicted measure of lead exposure via water during the years 2014-16 for every home in Flint, which can be used for public health studies in the years to come.

\subsection{Predictive Factors}

Feature importance with tree ensemble methods can be determined by the number of times the individual trees in the forest split on each feature.  We break down the results into the following categories.

\subsubsection{Home Value}

Various measures of a property's value were determined to be important by the model.  The top two features were consistently the value of the buildings and the value of the land. Additionally, land improvements and state assessed value were important.

\subsubsection{Demographics}

Demographic data from the census bureau was also important.  The model divided the city down by lines using age and race.  Some of the less important features that still contributed were whether homes had married parents and whether or not only English is spoken in the home.

\subsubsection{Property Age}

Finally, the age of the property was one of the most important features. This was visible in Figure $\ref{fig:loglead_by_year}$. Other values that were correlated to property age also appeared, such as the estimated age of the population and whether or not elderly people were present.

\subsection{Kaggle Challenge}

We initiated a Kaggle prediction challenge to improve our prediction accuracy. This was hosted by \texttt{https://inclass.kaggle.com/} and offered to people affiliated with the University of Michigan.
The contest involved a dataset with over 17,000 water tests from nearly 11,000 Flint homes de-identified. Along with the lead test results, some other de-identified features of the home and lead test, including property value, vacancy status, and time of test were provided.
During two months of competition, over 150 students and post-docs from various departments at the University of Michigan participated, submitting over 500 times in the process. The 1st, 2nd, and 3rd place winners had the opportunity to present their classifiers to the Michigan Data Science Team (MDST)\footnote{The authors are members of MDST, \url{http://midas.umich.edu/mdst/}.}. The result of the challenge was a small improvement to our initial classification models. The winning submission achieved this through ensembling XGBoost models with other classification models. However, the second and third winning solutions used a Random Forest model. We observed a high degree of variance between Random Forest submissions, in part due to the intrinsic uncertainty in the predictions.
Moreover, we learned the most significant improvements came through adding additional data, rather than hyperparameter tuning.

\begin{figure}[h]
\centering
\includegraphics[width=0.4\textwidth]{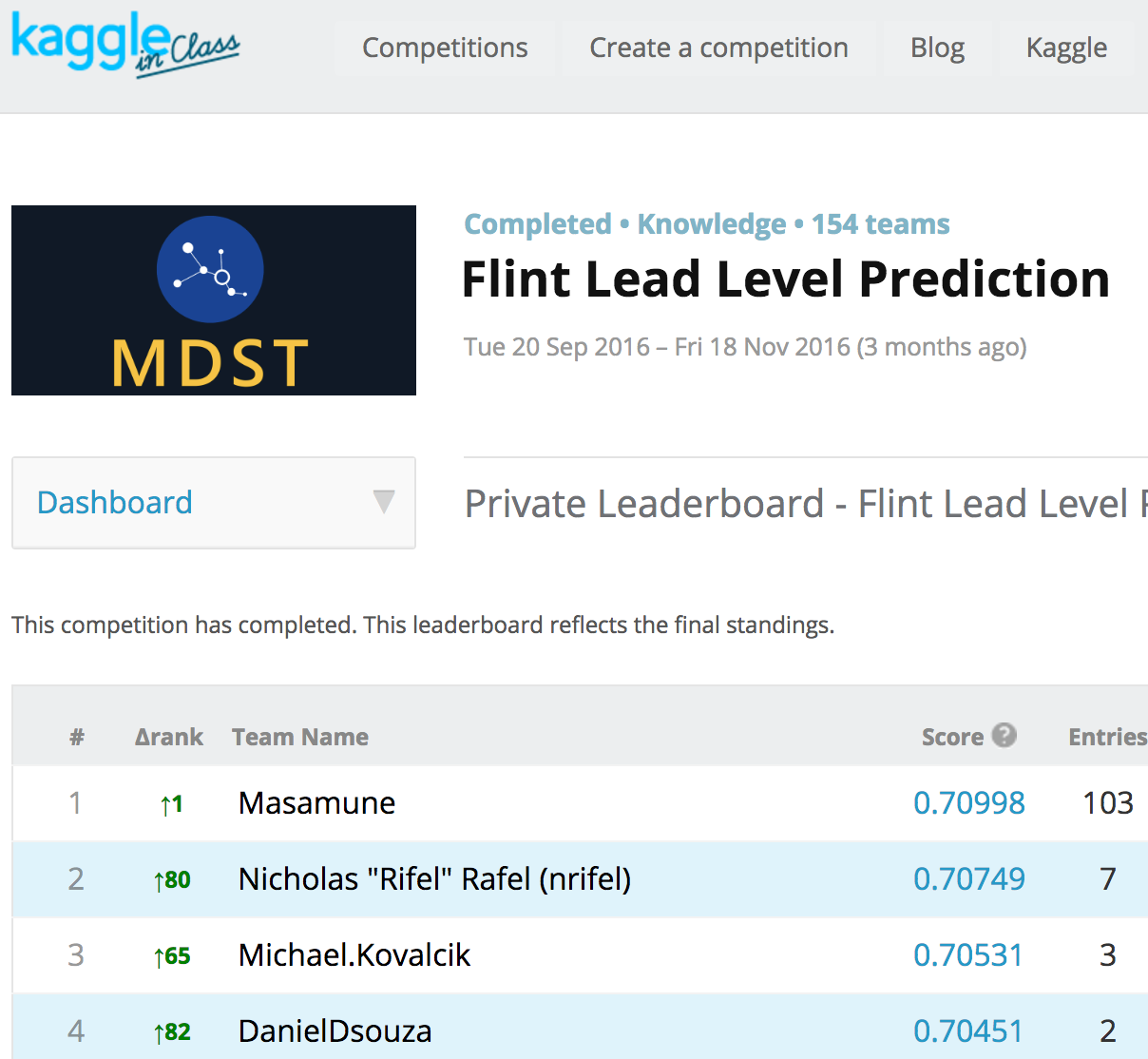}
\caption{The lead-level prediction problem was released as a UM internal prize-drive challenge. The competition was facilitated by the MSDT.}
\label{fig:kaggle_challenge}
\end{figure}

\subsection{The \texttt{MyWater-Flint} App}

\begin{figure}[h]
\centering
\vspace{0.1in}
\includegraphics[width=0.45\textwidth]{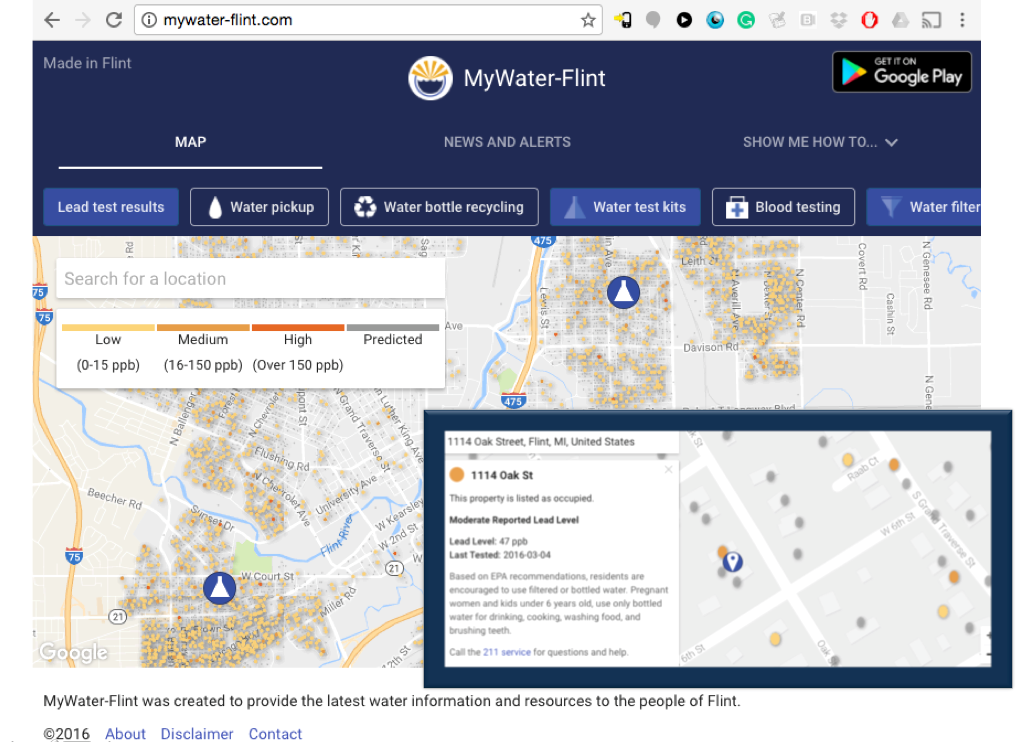}
\caption{Snapshot of the \texttt{Mywater-Flint} website.}
\label{fig:app_mywater_flint_combined}
\end{figure}

Related to our modeling efforts, we were involved in a project funded by \texttt{Google.org} to develop a mobile app and website for the city of Flint to help the community and government agencies manage the ongoing water crisis. Figure \ref{fig:app_mywater_flint_combined} shows a screenshot of the app. The app development was a collaboration between Professor Mark Allison at University of Michigan -- Flint, his students, and MDST, with support from \texttt{Google.org}.

The \texttt{Mywater-Flint} App \footnote{\url{http://www.mywater-flint.com/}}, uses the predictive model and features described earlier to identify homes at high, medium, and lows risk of lead contamination. The users are also able to do the following:

\begin{itemize}
\setlength{\parskip}{-3pt}
\item access a citywide map of where lead has been found in drinking water.
\item discover where service line workers have replaced infrastructure that connects. homes to the water main, and where they're currently working.
\item locate the nearest distribution centers for water and water filters.
\item find step-by-step instructions for water testing.
\item determine the likelihood that the water in a home or another location is contaminated, among other features.
\end{itemize}

\section{Characterizing Water Sample Submission Behavior}
\label{sec:behavior}

We find that of the 32,741 occupied homes, 10,998 submitted at least one water test. Investigating the predictive factors behind when and how often submissions occur can help us understand the submission behavior of residents. We study this behavior and investigate features which correlate with water test submission variables. 

\subsection{Predicting Which Homes Make Submissions} \label{sec:pred_fact_sub}

Despite the low cost of submitting a residential water test, a large majority of the properties in Flint have not submitted any tests. Many properties are simply vacant; these properties are discarded from the analysis in this section.  One hypothesis is that residents working long hours may not have the ability to conduct and deliver the test.  Another hypothesis is that some may not know where to obtain one.  In order to better understand why a property might make a submission, we employ several classifiers to predict whether a property has submitted.  Of these, we choose the best model according to accuracy of the classification.  We then calculate the feature importances to give insight into submission behavior. 

\subsubsection{Data Processing}

The dataset we use is the result of joining block level census data, city of flint parcel information, and the residential water testing dataset. Combined, the joined dataset contains 60 features and 32,741 rows where each row represents a parcel of land in Flint. As mentioned previously, vacant parcels are discarded. Then one-hot encoding is performed on all categorical features. The target variable is a binary where 0 means no submission and 1 means at least one submission. 

\begin{figure}[h]
\centering
\includegraphics[width=8.5cm]{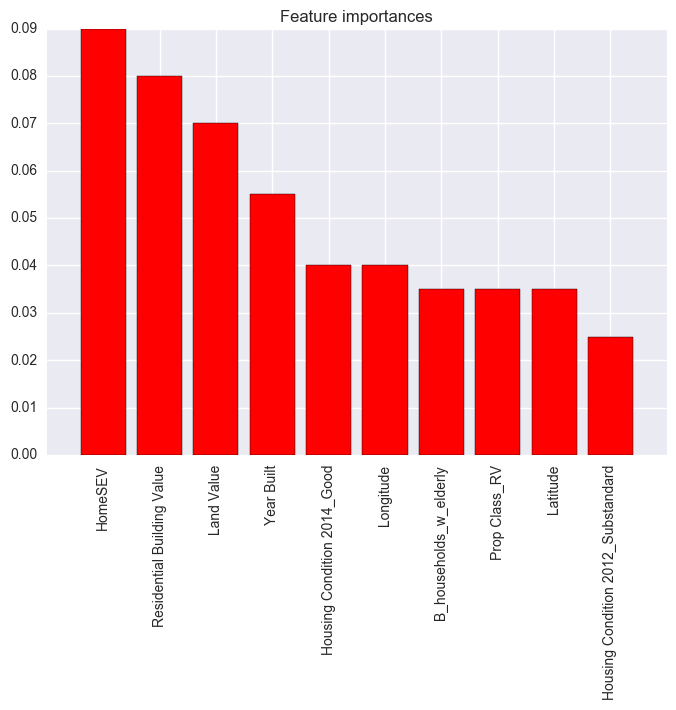}
\caption{This figure shows the 10 variables that an AdaBoost classifier deemed most important according to \emph{Gini importance} metric. The y-axis shows the the (normalized) total reduction of the criterion brought by that feature. Larger values indicate more important features.  Note that many of these features are related to parcel value.}
\label{fig:housing}
\end{figure}

\subsubsection{Model Selection and Training}

We use an AdaBoost classifier from the \texttt{scikit-learn} python package with \texttt{num\_estimators} and \texttt{learning\_rate} set to 200 and 0.2 respectively. We chose the AdaBoost model for it's robustness to overfitting and its consistent performance at this classification task when compared to logistic regression with L2 regularization. 

After training the model, we evaluate our performance using a 5-fold cross validation. The model consistently achieved recall accuracy of $0.65$ with a standard deviation of $\pm 0.03$, meaning the model correctly identified 64\% of the true positives in the cross-validation set. 

\subsubsection{Predictive Factors}

Of the the 60 features used in training the model, proxies for property value are consistently the most important features.
We calculate feature importances with the Gini importance metric \cite{breiman1984classification}.
Gini importance of a feature is computed by averages the Gini decreases for that feature over all trees \cite{treeratpituk2009disambiguating}.
See Figure $\ref{fig:housing}$ for a graph comparing the 10 most important features.

Table \ref{tab:most_pred_subm} describes the marginal distribution of some of the most predictive features. Parcels which submit more than one test are generally more valuable, as shown by increases in ``Residential Home Value'', ``HomeSEV'', and ``Parcel Acres''. We do not find that homes which are old, which would typically be at greatest risk for lead contamination, test less than other occupied properties. However, as illustrated in table \ref{tab:most_pred_subm_2}, the number of submissions from a property tends to increase with its value.

We find that of the various parcel, census, and infrastructure features considered by our models, features which describe the value of the parcel are more predictive than census demographic information.  However, the census data available to us are reported at the block level and may not be granular enough to inform the classifiers effectively.

\begin{table*}
\centering
\begin{tabular}{ |l|l|l|l|l|l| }
\hline
\multicolumn{6}{ |c| }{Parcel Features} \\
\hline
Attribute & Number of Submissions          & Q1           & Median       & Q3           & \% sample non-zero \\ \hline
\multirow{3}{*}{HomeSEV} & Zero            & \$7,500      & \$10,500     & \$14,000     & 43\% \\
                         & One             & \$8,600      & \$11,700     & \$16,500     & 67\% \\
                         & Two or more     & \$9,000      & \$12,400     & \$17,600     & 69\% \\ \hline
\multirow{3}{*}{Land Value} 
                         & Zero            & \$787        & \$1,697      & \$5,039      & 99\% \\
                         & One             & \$984        & \$2,652      & \$10,403     & 99\% \\
                         & Two or more     & \$1,074      & \$2,793      & \$15,984     & 99\% \\ \hline
\multirow{3}{*}{Residential Building Value} 
                         & Zero            & \$17,271     & \$32,891     & \$62,430     & 92\% \\
                         & One             & \$18,539     & \$36,294     & \$70,922     & 96\% \\
                         & Two or more     & \$19,338     & \$40,541     & \$80,478     & 96\% \\ \hline
\end{tabular}
\caption{This table gives the quartiles of the most predictive features. We find parcels with at least one submission are more valuable. Because the parcel data has some missing values, we include a column that indicates the number of non-zero values for the given category. }\label{tab:most_pred_subm}
\end{table*}

\begin{table*}
\centering
\begin{tabular}{ |l|l|l|l|l|l| }
\hline
\multicolumn{6}{ |c| }{Demographic Features} \\
\hline
Attribute        & Number of Submissions   & Year Built   & Q1           & Median        & Q3           \\ \hline
\multirow{3}{*}{Aggregate Income} 
                         & Zero            &  $>$1940       & \$37,416     & \$86,570      & \$127,476 \\
                         & One             &  $>$1940       & \$38,803     & \$87,589      & \$135,879 \\
                         & Two or more     &  $>$1940       & \$48,711     & \$100,411     & \$153,750 \\ \hline
\multirow{3}{*}{Aggregate Income} 
                         & Zero            &  $<$1940       & \$24,447     & \$62,301      & \$114,930      \\
                         & One             &  $<$1940       & \$34,783     & \$83,408      & \$143,736     \\
                         & Two or more     &  $<$1940       & \$34,783     & \$87,133      & \$214,893     \\ \hline
\end{tabular}
\caption{This table presents the quartiles of the household income for parcels who submitted zero, one, or more than one sample.  We also separte the homes into two groups based on property age.} \label{tab:most_pred_subm_2}
\end{table*}

\section{How Selection Bias affects Observed Lead Levels Over Time}

 Goovaerts (2017, \cite{goovaerts2017monitoring}) questioned the ``generalizability'' of sentinel sites and argued that sentinel sites are less representative than voluntary residential water test data.  However, the residential water test data could be biased due to the voluntary nature of the data collection process.  The analysis in this work shows that the important features in our water lead level prediction and water test submission submissions overlaps heavily.  One hypothesis is that water tests are less likely to be submitted from houses built before 1930, but those old houses are also those more likely to be suffering from high level lead exposure.  Thus to investigate whether the water lead level has improved over time quantitatively, we need to carefully correct the selection bias incurred by the data collection method \cite{goovaerts2017monitoring}.  We approach this problem by assigning correction weights on the residential test data when we calculated the quantile water lead level.  
 
\begin{figure}[h]
\centering
\includegraphics[width=0.48\textwidth]{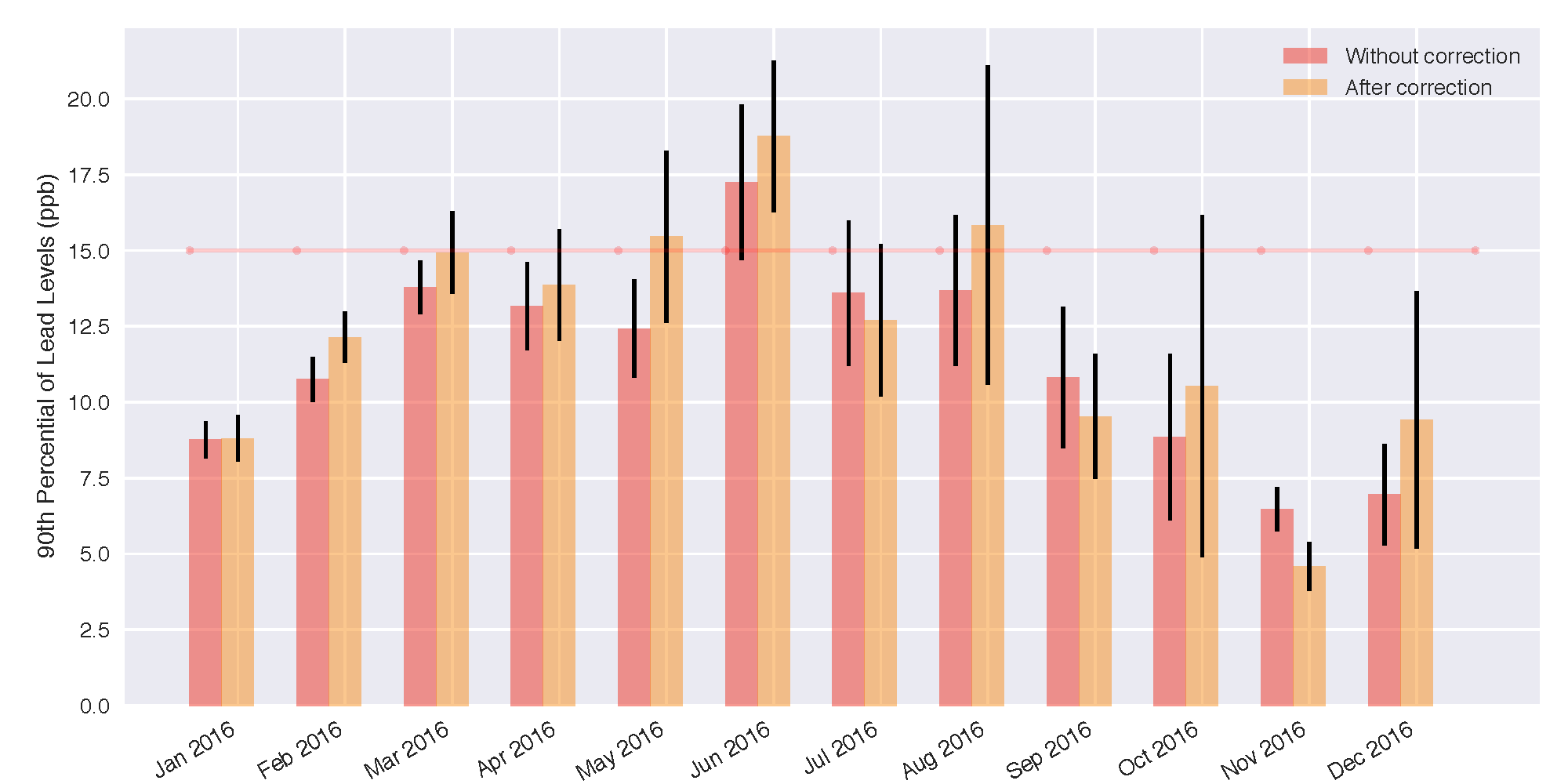}
\caption{Comparing the 90$^{\rm th}$ percentile of lead readings on voluntary testing data without/with the reweighting correction procedure for the selection bias. The error bar shows the standard deviation of the estimator by bootstrapping.}
\label{fig:nintieth}
\end{figure}

To get the weights, we take advantage of our predictive model for water test submission in section \ref{sec:pred_fact_sub}. The model provides the probability $p_i$ of each parcel $i$ submitting at least one water test sample. This probability is used as a proxy quantity for over-representation. Each observed sample should be given by a correction weight $w$ that is inversely proportional to  the (predicted) probability that it can be collected. Denote the set of collected samples in a given time period as $\mathrm{S}$, $w_i = \frac{\sum_{i\in \mathrm{S}}p_i}{p_i}.$

For any water test that couldn't match government parcel records, we assign an median weight and then we normalize each month's total weight to 1. After the weighting procedure, we examine the water lead level improvement over time. We note that that despite the lack of sampling strategy, the correction doesn't change the conclusion that over the whole year of 2016, the water lead level dropped after reaching the highest level at about May. Goovaerts (2017) adapted a weighted average of stratum-specific rates to estimate the effect of sampling bias and concluded that voluntary testing capture the main characteristics of Flint properties much more closely than the sentinel program \cite{goovaerts2017monitoring}. Though they are using a different approach their findings are consistent with findings in this paper.

After the bias correction, the 90$^{\rm th}$ percentile estimate of water lead level in some months increase by a small amount, which is in favor of our hypothesis that the selection bias mostly results from the lack of submission from the old houses most affected by the crisis. This trend has been noticed elsewhere \cite{goovaerts2017monitoring}. Modern correction techniques may be able to provide better insights, which is beyond the scope of this work.

\section{Conclusions}

The lead contaminating Flint's water systems poses a serious health risk for all of the city's residents.
There are two major challenges with assessing water contamination using samples tested for lead. The first is that the observed distribution of lead levels in water fat tailed and highly skewed: the 95$^{\rm th}$ percentile of Flint's lead readings is 28 ppb, the 99$^{\rm th}$ percentile is 180 ppb, and the 99.9$^{\rm th}$ percentile is over 2,100 ppb. The second challenge is that measuring lead contamination is a highly noisy process.

We collaborated with the City of Flint and the Michigan Department of Environmental Quality to acquire data and joined these data with existing public data.
We used these data to build a predictive model to predict which homes are more likely at risk of high lead contamination. This model is employed in predictions shown on the \texttt{MyWater-Flint} app and website.  We identified features which are strong predictors of high lead levels and found that a number of factors, not just the composition of service lines, are important to consider in addressing the crisis. 
Knowing these risk factors can help policy makers and community members better allocate limited resources and prioritize action in this time of need.

Our lead predictions may also have future value. By establishing each home's chance of having had high lead during 2014-16 crisis, this work provides a proxy for lead exposure to be used studies tracking health outcomes for Flint residents in years to come.

This work is ongoing and serves as a model for university-community partnerships and for data-driven public policy decision making.

\section*{Acknowledgments}

This work was supported by \texttt{Google.org} and the National Science Foundation, grants IIS 1453304 and IIS 1421391. 
This work would not have happened without the support of the broader Michigan Data Science Team, including Jonathan Stroud and many
others. The authors recognize the support of Michigan Institute for
Data Science (MIDAS) and computational support from NVIDIA.

\bibliographystyle{ACM-Reference-Format}
\balance
\bibliography{sigproc} 

\end{document}